# A Novel Multiple Ensemble Learning Models Based on Different Datasets for Software Defect Prediction

**Ali Nawaz**
Department of Computer and Software Engineering, CEME,
National University of Sciences and Technology, Islamabad, Pakistan
anawaz.cse19ceme@ce.ceme.edu.pk

**Attique Ur Rehman**
Department of Computer and Software Engineering, CEME,
National University of Sciences and Technology, Islamabad, Pakistan
aurehman.cse19ceme@ce.ceme.edu.pk

**Muhammad Abbas**
Department of Computer and Software Engineering, CEME,
National University of Sciences and Technology, Islamabad, Pakistan
m.abbas@ceme.nust.edu.pk

**Abstract:**

. Software testing is one of the important ways to ensure the quality of software. It is found that testing cost more than 50% of overall project cost. Effective and efficient software testing utilizes the minimum resources of software. Therefore, it is important to construct the procedure which is not only able to perform the efficient testing but also minimizes the utilization of project resources. The goal of software testing is to find maximum defects in the software system. More the defects found in the software ensure more efficiency is the software testing Different techniques have been proposed to detect the defects in software and to utilize the resources and achieve good results. As world is continuously moving toward data driven approach for making important decision. Therefore, in this research paper we performed the machine learning analysis on the publicly available datasets and tried to achieve the maximum accuracy. The major focus of the paper is to apply different machine learning techniques on the datasets and find out which technique produce efficient result. Particularly, we proposed an ensemble learning models and perform comparative analysis among KNN, Decision tree, SVM and Naïve Bayes on different datasets and it is demonstrated that performance of Ensemble method is more than other methods in term of accuracy, precision, recall and F1-score. The classification accuracy of ensemble model trained on CM1 is 98.56%, classification accuracy of ensemble model trained on KM2 is 98.18% similarly, the classification accuracy of ensemble learning model trained on PC1 is 99.27%. This reveals that Ensemble is more efficient method for making the defect prediction as compared other techniques.

*Keyword:* **Software Quality Engineering, Software testing, Machine learning, Supervised learning**

## 1. Introduction
The success of any software completely depends on proper software development process and testing is the important phase of software development life cycle. It is important step to ensure quality in the software as minute defect in the software effects the later stages of software development tremendously. Software testing consumes more than 50% of the overall development cost [1] and the effort required by the software testing is approximately 40-60% of the overall development process [2]. Therefore, software testing needs to manage efficiently so that resources would be effectively utilize. Software testing can be performed either manually or automated [2lit]. Manual testing is time consuming as well as inaccurate due to involvement of human being as compared to automated testing, which is more accurate and time sufficient. The goal of software testing is to find maximum number of defects in the software [7]. Software defect prediction is the process to identify the defects in the software. Early finding of defects not only effects the quality of software but also helps the effective utilization of resources.

Machine learning is the widely used technique to predict or find defects in the software [8], [9]. From last few decades the applicability of machine learning in the real-world problems rises due to availability of huge amount of labelled data. Machine learning is the ability of computer to learn from data [2]. Machine learning is classified into three categories 1) Supervised learning 2) Unsupervised learning 3) Semi-supervised learning. In supervised learning, there is both features and labels while in unsupervised learning there is feature only and in semi-supervised learning there is small amount of labelled data and huge amount of unlabeled data. The application of supervised learning is more than unsupervised and semi supervised learning. Unsupervised learning is also widely used for software defect detection [3], [4], [5], [6]. Supervised learning is further categorized into classification and regression. In regression, the labels are continuous variables while the labels in classification are discrete variables. In this paper we are dealing with classification task as datasets are labels with discrete variables.

This paper proposed the ensemble leaning technique for increasing the accuracy of defect prediction. The datasets used in the experiment is the open source data of PROMISE repository [7] CM1, KC1, PC1. The result of the proposed technique is compared with few prominent machine learning algorithms such as K nearest neighbors (KNN), support vector machine (SVM), decision tree (DT) and it is shown that results of proposed ensemble learning method is effective and sound. The evaluation metrics used for comparison of results are precision, recall, accuracy, F1-score and ROC curve.

The main contributions of the papers are summarized as;

- Proposed ensemble learning model for the defect prediction
- Detailed comparison of proposed model with prominent models of machine learning
- Achieve defect prediction accuracy of maximum 99%.

The rest of the research paper is organized as follows. The next section will review some literature. The proposed methodology is described in Section III and the corresponding experimental results are shown in Section IV. The final section concludes the research paper.

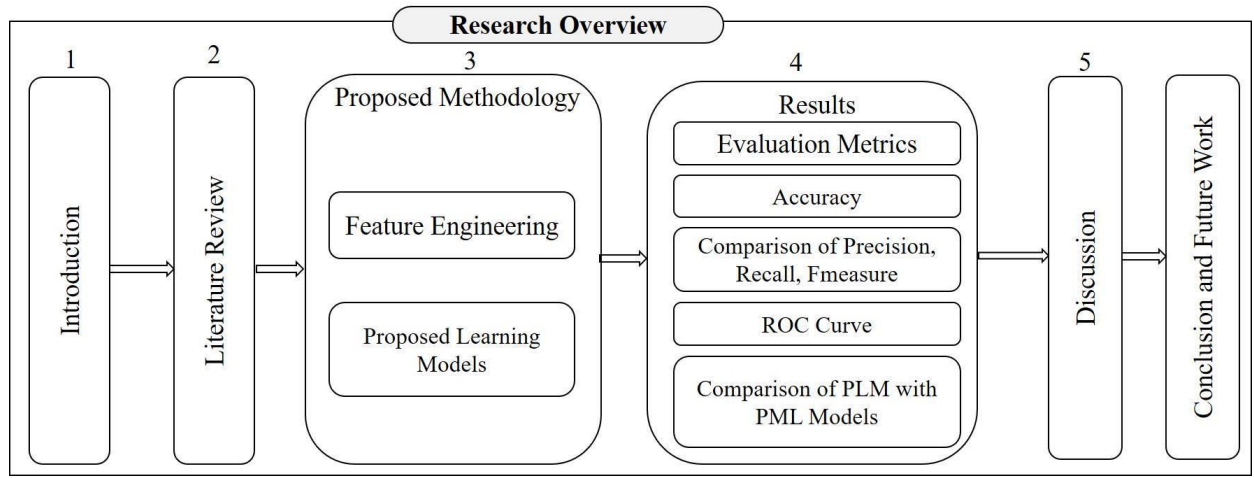

**Figure 1:** Research Methodology Overview

## 2. Literature Review:

According to [10], software testing is the integral part pf software development life cycle. The overall success of software relies on the testing phase. A large number of machine learning techniques are proposed for efficient defect detection in the software. C. Manjula et al. [11] proposed a hybrid machine learning approach in which Genetic algorithm is presented to improve fitness function and for the better optimization of the features then the optimized features are processed through decision tree (DT). The performance is compared with ID3 based decision tree and it is proven that proposed hybrid approach achieve better results. The proposed approach successfully addresses the performance challenge.

I Laradji et al. [12] proposed ensemble learning technique with more emphasis on feature engineering. The proposed method combines ensemble learning and efficient feature selection to address the robustness problem of previous defect prediction techniques. The proposed method also presents average probability ensemble (APE) which is ensemble of seven machine learning models and reveals that the efficiency of APE is improved over other machine learning models such as weighted SVM and random forests. It is found that the APE combines with greedy forward selection produce better results for PC2, PC4 and MC1 datasets.

O Arar et al. [13] proposed a hybrid technique for defect prediction in which Artificial Neural Network (ANN) is use for making prediction and the weight of ANN is optimized through Artificial Bee Colony (ABC) algorithm. The performance of proposed method is compared with Naïve Bayes, Random Forest, C4.5, Immunos and AIRSand algorithms and found that the performance of proposed technique and random forest is equal on KC1 datasets and produces high accuracy on KC2 and CM1 datasets. However, performance is not good on PC1 and JM1 datasets as compared to other techniques. The author suggests that more focus on the feature engineering may reveal better results.

M Siers et al. [14] proposed ensemble method for classification of defect. The proposed method is the ensemble of decision tree called as CSForest. For minimizing the classification cost a cost sensitive coting technique called CSVoting is proposed. The evaluation of proposed technique is performed on six prominent classifiers C4.5, SVM, SysFor+Voting1, SysFor+Voting2,

CSC+C4.5, CSTree and six publicly available datasets. The proposed method shows that the lower prediction cost is achieved by combining CSForest and CSVoting.

P Singh et al. [15] performed an analysis of prominent machine learning i.e ANN, PSO (Particle Swarm Optimization), DT, NB and LC (Linear classifier) and evaluated on seven PROMISE [16] datasets. These algorithms are analyzed by using KEEL tools and validated using k-fold cross validated technique. The results of the analysis reveal that LC has highest defect prediction accuracy in four out of seven datasets then followed by Naïve Bayesian, Decision Tree and Neural Network having second highest accuracy in one dataset therefore LC is more dominant over other classifiers.

A Hammouri et al. [17] performed the comparative analysis of Naïve Bayes (NB), Decision Tree (DT) and Artificial Neural Networks (ANNs) classifiers for software defect prediction. the performance of the proposed method is evaluated on three publicly available datasets and performance metrics used are accuracy, precision, recall, F-measure and ROC curve. The results reveal that the average accuracy of DT is more than the ANN and NB. Experimental results also reveals that ML algorithms provide better defect prediction accuracy than other algorithms such as linear AR and POWM.

R Jayanthi et al. [18] proposed a software defect prediction approach in which firstly they perform dimensionality reduction via Principle Component Analysis (PCA), a most famous dimensionality reduction approach and PCA is further improved computing maximum likelihood of error estimation of PCA. After dimensionality reduction, ANN is applied to perform the prediction of defects. The performance is evaluated on four publicly available datasets i.e KC1, PC3, PC4 and JM1 and evaluation metrics are precision, recall, classification accuracy etc. Experimental results reveal that by using the proper feature engineering the time and space complexity is reduced with effecting accuracy of defect prediction

C. Manjula et al. [19] proposed a hybrid machine learning approach in which Genetic algorithm is presented to improve fitness function and for the better optimization of the features then the optimized features are processed through Deep Neural Network (DNN). The performance is compared with Naïve Bayes, SVM, Decision Tree, KNN and other several ML algorithms and it is proven that proposed hybrid approach achieve better results. The proposed approach successfully addresses the performance challenge. The classification accuracy of proposed approach is 97.82% on KC1 dataset, 97.59% on CM1 dataset, 97.96% for PC3 dataset and 98.0% for PC4 dataset which is better than previous techniques.

A Iqbal et al. [20] proposed a feature selection based ensemble method the defect prediction. The proposed is implemented by both with and without feature selection. The performance is evaluated on four publicly available datasets of PROMISE and evaluation metrics are Precision, Recall, Fmeasure, Accuracy, MCC and ROC. The performance of proposed method is compare with Naïve Bayes (NB), Multi-Layer Perceptron (MLP), Radial Basis Function (RBF), Support Vector Machine (SVM), K Nearest Neighbor (KNN), kStar (K*), One Rule (OneR), PART, Decision Tree (DT), and Random Forest (RF) and reveals that proposed method generate better results.

**Table 1:** Summary of Literature Review

| Ref. | Proposed Method | Datasets | Evaluation metrics and results | Advantages |
|---|---|---|---|---|
| [11] | Hybrid Genetic algorithm and DT ensemble is proposed for defect prediction | PROMISE dataset (PC3, PC4, KC3) | (Accuracy) PC3 = 91.68%, PC4 = 92.09%, KC3 = 93.36% | Performance of defect detection is successfully increasing |
| [12] | Ensemble learning with better feature selection reduces the robustness in the defect prediction | PC2, PC4 and MC1 datasets | AUC is closed to 1.0 for all datasets | Addresses the robustness limitation of previous prediction methods |
| [13] | A hybrid of ANN and ABC classification algorithm is proposed for defect prediction | PC1, JM1, KC1, KC2 and CM | KC1 = 0.80, KC2 = 0.85, CM1 = 0.77, PC1 = 0.82, JM1 = 0.71 | Performance is better than few methods. |
| [14] | Ensemble CSForest and CSVoting technique to minimize the cost sensitivity of classification algorithm | MC2, PC1, KC1, PC3, MC1 and PC2 | (Precision) MC2 = 0.446 KC10 = 0.272, PC3 = 0.455, PC3 = 0.392, MC1 = 0.639, PC2 = 0 (Recall) MC2= 0.375 KC1 = 0.101, PC3 = 0.168, PC3 = 0.179, MC1 = 0.093, PC2 = 0 | lower prediction cost is achieved |
| [15] | Compare the performance of ANN, PSO, DT, NB and LC | CM1, JM1, KC1, KC2, PC1, AT, KC1CL | (LC Accuracy) CM1 = 87.95%, JM1 = 80.84%, KC1 = 84.49%, KC2 = 82.73%, PC1 = 93.59%, AT = 90.76%, KC1CL = 66.80% | |
| [17] | Compare the defect prediction performance of ANN, DT and NB | DS1, DS2 and DS3 | (Accuracy) NB = 93%, DT = 97, ANN = 95% | ML is better defect prediction technique than other techniques |
| [18] | After applying PCA, the performance of PCA is improved by computing maximum likelihood error estimation of PCA and finally ANN is applied for defect prediction | KC1, PC3, PC4 and JM1 | (Accuracy) JM1 = 90.93%, PC4 = 93.64%, KC1 = 86.91%, JM1 = 83.03% | Reduces time and the space required visualize the data by reducing the data to low dimensions |
| [19] | Hybrid Genetic algorithm and DNN for defect detection | KC1, CM1, PC3 and PC4 | (Accuracy) KC1 = 97.82%, CM1 = 97.59%, PC3 = 97.96% PC4 = 98.0% | Highest accuracy over current defect prediction techniques |
| [20] | feature selection-based ensemble classification framework | CM1, JM1, KC1, KC3, MC1, MC2, MW1, PC1, PC2, PC3, PC4 and PC5 | (Highest Accuracy) JM1 = 74%, KC3 = 86%, MC1 = 88%, PC4 = 95%, PC5 = 80% | |

## 3. Proposed Methodology
### 3.1 Feature Engineering

In this section, a detailed analysis of datasets and basic feature processing performed of the datasets are also explained.

#### 3.1.1 Feature Description

The datasets used in this paper is the open source data which is created and distributed by the NASA Data Metric Program [16]. The datasets are available in different versions and for our experimentation we are just using CM1, KC2 and PC1. The CM1 dataset is composed of 498 instances and 21 attributes similarly, KC2 is composed of 522 instances and 21 attributes and PC1 composed of 1109 instances and 21 attributes. The complete description of attributes is given in Table 2.

**Table 2.** The attributes and their explanation

| S. No. | Attributes | Explanation |
|---|---|---|
| 1. | loc | Total number of lines of code (McCabe's) |
| 2. | v(g) | Cyclometric complexity analysis (McCabe's) |
| 3. | ev(g) | Essential complexity (McCabe's) |
| 4. | iv(g) | Design complexity (McCabe's) |
| 5. | n | Total number of operator and operands |
| 6. | v | Halstead Volume |
| 7. | l | Halstead program length |
| 8. | d | Halstead difficulty |
| 9. | i | Halstead intelligence |
| 10. | e | Halstead measurement effort |
| 11. | b | Halstead estimation effort |
| 12. | t | Halstead time estimator |
| 13. | loCode | Halstead total number line of code |
| 14. | loComment | Halstead total number line of comment |
| 15. | loBlank | Halstead total number of blank lines |
| 16. | loCodeandComment | Halstead total number of lines and comments |
| 17. | uniq_op | Total number of unique operators |
| 18. | uniq_opnd | Total number of unique operands |
| 19. | total_op | Total number of operators |
| 20. | total_opnd | Total number of operands |
| 21. | branchCount | Total number of branches in each module |

### 3.1.2 Feature Selection

The main step of feature engineering is to select feature relevant [21] to the domain of problem we are going to solve and ignore the irrelevant features. All the features of the datasets in Table 2 are important and support the defect prediction task.

### 3.1.3 Handling class imbalance

During the analysis of the datasets it is observed that data is not balanced in some classes there are more instances and other have less instances. The Cm1 has approximately 90% false (not-defect) values and 10% true (defect) values. Similarly, KC2 has 20% true (defect) values and 80% (non-defect) values as compared to PC1 which has 93%(defect) and 7% (non-defect) values. So, the problem of class imbalanced is solved by sample Bootstrapping [22] and the number of observations used for sampling is 7. Bootstrapping is also known as Bootstrap aggregation is a random sampling with replacement method for creating the random samples of the data.

### 3.2 Proposed learning models
### 3.2.1 Proposed learning model trained on
The proposed ensemble learning model trained on CM1 is shown in Figure 2 which shows that after performing basic feature preprocessing an ensemble learning is applied which is ensemble of Classification by Regression [23] and KNN model [24]. After the ensemble method, KNN is applied again for achieving better classification accuracy. The classification by regression model has multiple subprocesses and subprocesses has operators that is trained on regression model then the operator of the regression model is trained by classification model.

KNN acronym of K nearest neighbor is the prominent classification model. It finds the distance between test point and every point on the training data, then find the k nearest neighbor between the points.

### 3.2.2 Proposed learning model for PC1 and KC2
The proposed model for PC1 and KC2 is shown in Figure 3 and 4 respectively which present that after performing basis feature engineering an ensemble model is applied which is ensemble of Bagging and KNN. After ensemble learning method a KNN model is applied again for betterment of the accuracy and is observed that approximately 99% accuracy is achieved.

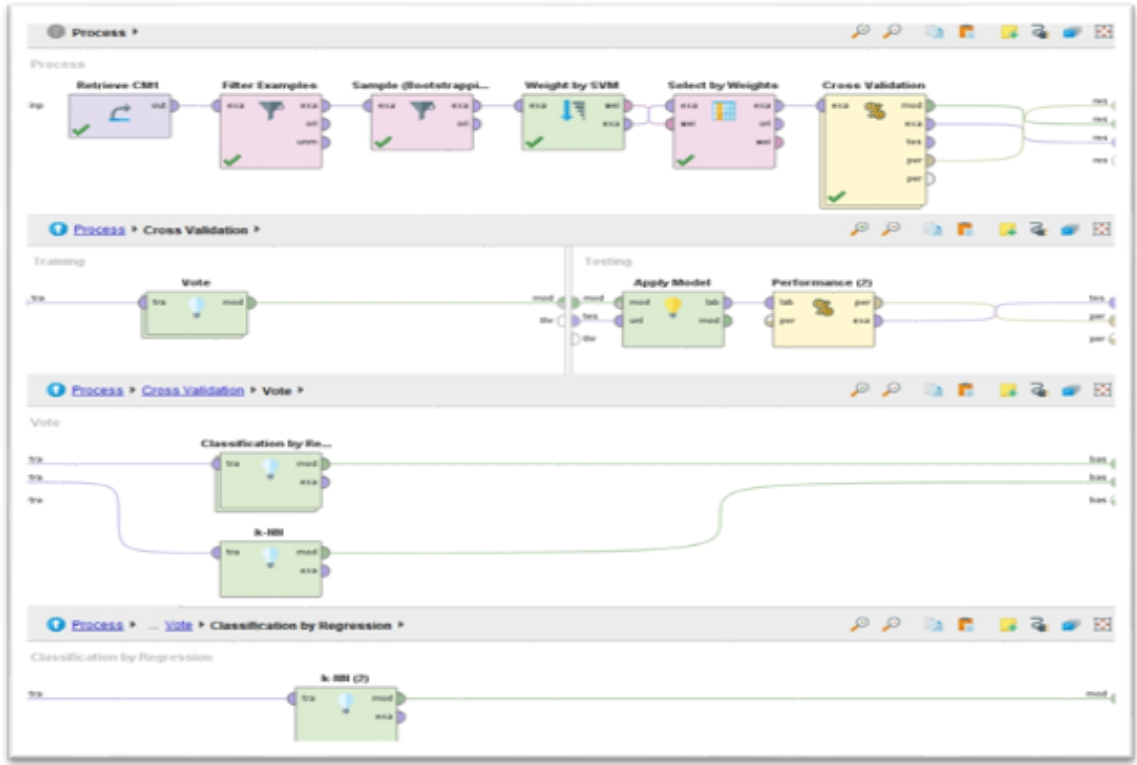

**Figure 2:** Proposed Ensemble learning model of CM1

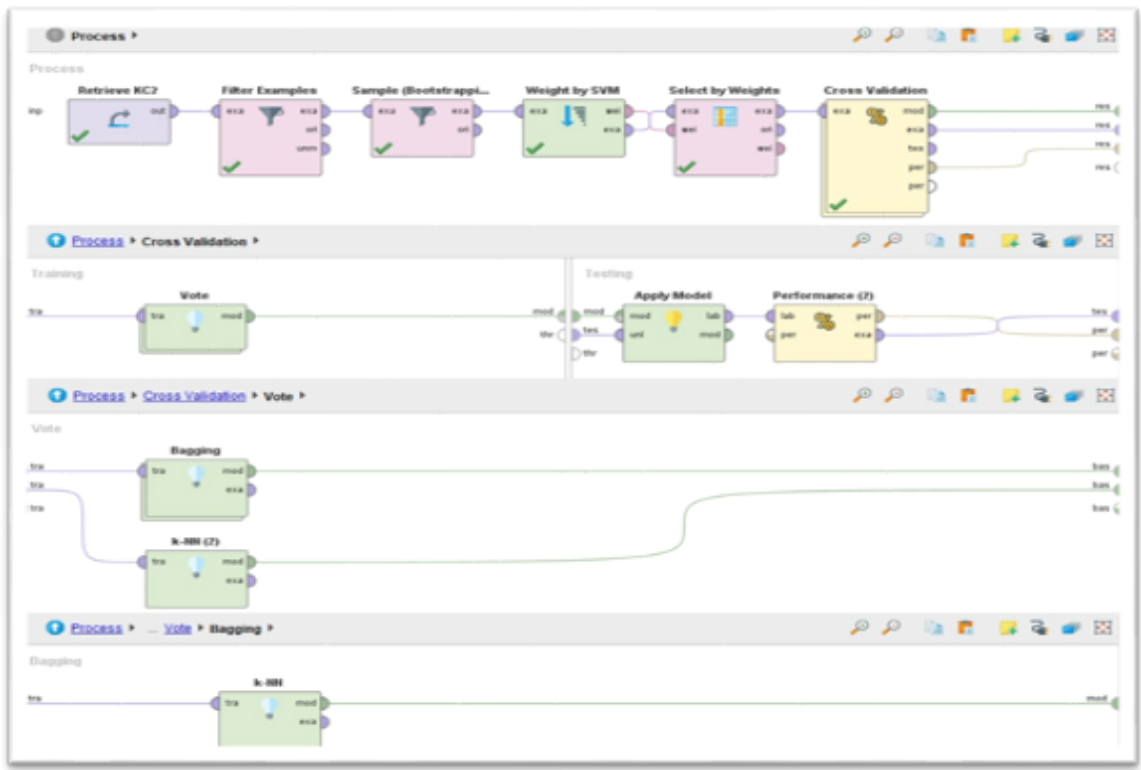

**Figure 3:** Proposed Ensemble learning model trained on

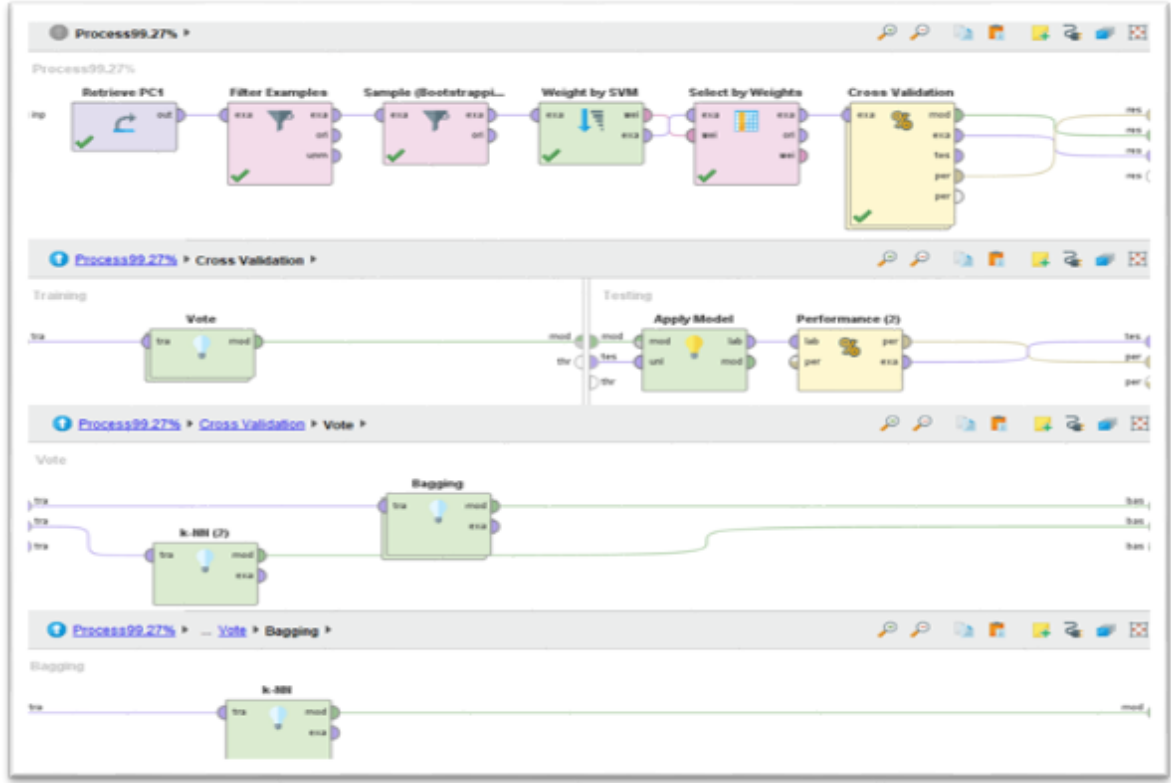

**Figure 4:** Proposed Ensemble learning model of ensemble learning model trained on PC1

## 4. Results
### 4.1 Evaluation metrics

The most important evaluation metrics [25] we are using in the experiment are described in Equation (1):

**Table 3:** Classification metrics

| Predicted Value | | Actual Value | |
|---|---|---|---|
| | | Positive | Negative |
| | Positive | True Positive (TP) | False Positive (FP) |
| | Negative | False Negative (FN) | True Negative (TN) |

Accuracy = TP + TN / TP + TN + FP + FN

Precision = TP / TP + FP          Equation (1)

Recall = TP / TP + FN

F1-score = 2 x Precision x Recall / Precision + Recall

### 4.2 Accuracy of train and test datasets

It is observed that accuracy of ensemble learning model trained on CM1 is 98.56% similarly, accuracy of model trained on KC2 is 98.17% and accuracy of ensemble learning model trained on PC1 is 99.27%. As it reviewed that the maximum percentage of accuracies achieved previously was 97.59% [19] on CM1. Our experimentation results demonstrate that ensemble learning produce more better results than simple learning models [26].

### 4.3 Comparison of Precision, Recall, F1-Score of proposed learning model

The formulae to calculate precision, recall and F1-score are given in Equation (1). Experimental results demonstrate that our proposed ensemble learning method achieved more accuracy than the previous methods. The comparison of results of precision, recall and F1-score is demonstrated in Table 4, 5, 6.

### 4.4 ROC curve

ROC curve acronym of receiver operating characteristics curve is the machine learning evaluation tool for analyzing the behavior of different classifiers at different threshold [27]. It is the graphical representation between False Positive Rate (FPR) and True Positive Rate (TPR). The ROC shows that Ensemble learning model is achieve more accuracy than previous proposed model. The ROC curves of models are shown in Figure 5.

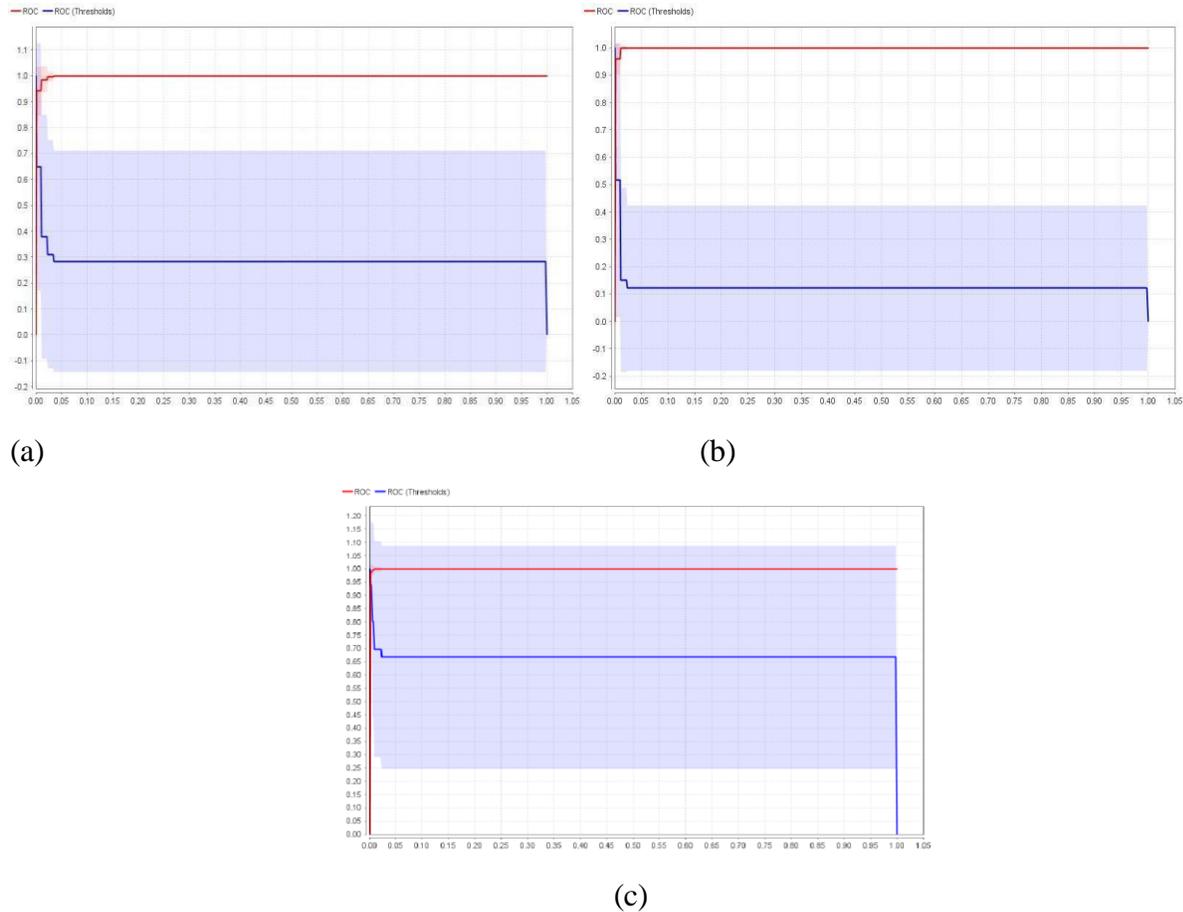

(a)  (b)

(c)

**Figure 5:** ROC of proposed ensemble learning model **(a)** ROC on CM1 **(b)** ROC on KC2 **(c)** ROC on PC1. In probabilistic model when output is greater than certain threshold identified as positive, then the element of confusion matrix totally depends on the threshold [28].

### 4.5 Comparison of Proposed learning model with prominent Machine learning model

In this section, the comparison between proposed ensemble learning model is compared with prominent machine such as SVM, KNN, Decision Tree and Random Forest [29], [30], [31], [32]. The comparison is performed on the basis of train and test accuracy, precision, recall, F-score which is shown in Table 4, 5, 6. The accuracy achieved on CM1 by applying SVM is 89.87%, KNN is 95.41% and accuracy of DT is 94.26% and on Random Forest accuracy is 92.54%. Similarly, accuracy achieved on KC2 by applying SVM is 83.85%, KNN is 89.92% and accuracy of DT is 88.59% and on Random Forest accuracy is 92.43%. Also, the accuracy achieved on PC1 by applying SVM is 92.66%, KNN is 98.5% and accuracy of DT is 94.89% and on Random Forest accuracy is 95.36%. The result of comparison demonstrate that ensemble learning model is more suitable model for this type of datasets for software defect prediction.

**Table 4:** Comparative analysis of prominent ML models w.r.t CM1 Dataset

| Model | Accuracy | Precision | Recall | F-score |
|---|---|---|---|---|
| SVM | 89.87% | 89.95% | 99.89% | Unknown |
| KNN | 94.41% | 99.36% | 99.57% | 95.41% |
| Decision Tree | 94.26% | 94.31% | 99.68% | 57.75% |
| Random Forest | 92.54% | 92.34% | 100% | 40.91% |
| Proposed Ensemble | 98.56% | 98.8% | 99.62% | 99.62% |

**Table 5:** Comparative analysis of prominent ML models w.r.t KC2 Dataset

| Model | Accuracy | Precision | Recall | F-score |
|---|---|---|---|---|
| SVM | 83.85% | 84.41% | 97.7% | 43.81% |
| KNN | 96.26% | 95.51% | 100% | 89.93% |
| Decision Tree | 88.59% | 92.69% | 93% | 71.91% |
| Random Forest | 92.43% | 91.26% | 100% | 77.87% |
| Proposed Ensemble | 98.17% | 97.75% | 100% | 95.22% |

**Table 6:** Comparative analysis of prominent ML models w.r.t PC1 Dataset

| Model | Accuracy | Precision | Recall | F-score |
|---|---|---|---|---|
| SVM | 92.66% | 92.66% | 100% | Unknown |
| KNN | 98.5% | 98.68% | 99.72% | 88.22% |
| Decision Tree | 94.89% | 94.86% | 99.91% | 45.16% |
| Random Forest | 95.36% | 95.22% | 100% | 56.10% |
| Proposed Ensemble | 99.27% | 99.86% | 99.35% | 95.13% |

## 5. Discussion

Software testing consumes more than 50% resources of overall software development process. Defect detection is one of the important activities of software testing. Early detection of defects reduces the consumption of resources. There are many techniques proposed for prediction of defects. Machine learning is widely used technique for the detection of software defects and produces efficient accuracy as well. Particularly, Machine learning algorithm i.e. SVM, KNN, DT, RF, NN, DNN, GA and their different variation are widely used for defect prediction. In this paper, we proposed three Ensemble learning models and trained on three different datasets i.e. CM1, KC2, PC1. The datasets are publicly available in promise repository which is created by NASA Data Metric Program (DMP) and composed of 21 continuous variables and 1 label with two classes i.e. Defect or Not Defect.

Before applying machine learning model, the class imbalanced problem which is the common problem in machine learning is handled by using sample with replacement technique called Bootstrapping. It is observed that ensemble learning model trained on CM1 produces 98.56% accuracy and ensemble learning model trained on KC2 produces 98.18% accuracy similarly ensemble learning model of ensemble learning model trained on PC1 produces accuracy of 99.27%. The proposed model is compared with prominent machine learning model i.e. SVM, KNN, DT and RF and it is observed that proposed model produces high result than all other models. The comparison between models are performed on the basis of evaluation metrics i.e. precision, recall, F1-score and accuracy. The experimental results demonstrate that ensemble learning models are efficient for software defect prediction over other machine learning techniques.

## 6. Conclusion and Future Work

It is concluded that software testing is the most important way to ensure quality in the software system. Defect detection and reduction is the most important part of software testing. Early detection of software defects reduces the consumption of resources. To detect the defect in the software various techniques has been proposed and achieved sufficient results. Among that techniques machine learning techniques produces more sufficient results. In this paper, we proposed ensemble learning model for predicting the defects in the software. The proposed model is trained on the Promise datasets. The performance of proposed model in term of accuracy metrics i.e. accuracy, precision, recall and F1-score is compared with other prominent machine learning algorithms i.e. SVM, KNN, Decision Tree and Random Forest and it is observed that proposed ensemble learning model produces better results. Classification accuracy of ensemble model trained on CM1 is 98.56%, Classification accuracy of ensemble model trained on KC2 is 98.18% and classification accuracy on PC1 is 99.27%. The proposed ensemble learning model can help Software Engineer for efficient detection and prediction of software defects earlier. The future work will focus on proposing general ensemble learning model for all the available datasets of promise repository with better accuracy of approximately 97% or above.

Intelligence, Networking and Parallel/Distributed Computing (SNPD), pp. 399 -406. IEEE, 2018.